\documentclass[10pt,twocolumn,letterpaper]{article}

\usepackage{wacv}
\usepackage{times}
\usepackage{epsfig}
\usepackage{graphicx}
\usepackage{amsmath}
\usepackage{amssymb}

\usepackage{color}

\usepackage{array}
\usepackage{multirow}
\usepackage{hhline}
\usepackage{booktabs}
\usepackage{rotating}

\usepackage{epstopdf}
\usepackage{xparse}
\usepackage{lscape}

\usepackage[pagebackref=true,breaklinks=true,letterpaper=true,colorlinks,bookmarks=false]{hyperref}

\wacvfinalcopy 

\def\wacvPaperID{23} 

\ifwacvfinal\fi
\begin{document}

\title{Gradient Boundary Histograms for Action Recognition}

\author{Feng Shi,  Robert Lagani\`{e}re and Emil Petriu\\
School of Electrical Engineering and Computer Science\\
University of Ottawa, Ottawa, On, Canada\\
{\tt\small \{fshi098, laganier, petriu\}@eecs.uottawa.ca}
}

\maketitle


\begin{abstract}
This paper introduces a high efficient local spatio-temporal descriptor, called gradient boundary histograms (GBH).
The proposed GBH descriptor is built on simple spatio-temporal gradients, which are 
fast to compute. We demonstrate that it can better represent local structure and motion 
than other gradient-based descriptors, and significantly outperforms them on large realistic datasets.
A comprehensive evaluation shows that the recognition accuracy is preserved
while the spatial resolution is greatly reduced, which yields both high efficiency and low
memory usage.

\end{abstract}

\section{Introduction}
Recent studies in human action recognition have achieved remarkable performance. Over the years,
 the progress has shown almost perfect results 
on atomic actions captured under controlled settings. As a result, the research community  now 
focuses on realistic datasets with relatively large number of classes, and  very good
results are reported \cite{Shi2013a,Zhu2013,wang2013b}.
However, much effort has been invested in improving the recognition accuracy with less 
consideration in efficiency. For example, 
the state-of-the-art approach  \cite{wang2013b} combines both SURF feature matches and
dense matches from optical flows to estimate the homography for camera motion compensation.
To further improve the robustness, it also uses a sophisticated  human detector as well as human tracking to
remove matches from the foreground human regions. All these techniques add  more complexity to
the already high cost trajectories-based method, which is built on dense optical flow.

The  low-level local  spatio-temporal features and 
 bag-of-features(BoF) \cite{Shi2013a,Shi2013,Wang2011} representation or alternative Fisher 
 vector encoding \cite{wang2013b,Oneata2013} can
 achieve  good performance for action recognition on realistic datasets.
A key factor for high performance is the local descriptors, which 
should include both local structure and motion information. Among all descriptors, 
 MBH  outperforms other descriptors by encoding motion boundary and suppressing camera motion.
 However, to build up MBH,  dense optical flows are computed for  consecutive frames.
 Dense optical flow is expensive to compute considering the large amount of video data to be processed.
 In addition, it includes two descriptors,  MBHx and MBHy, which add to the dimensionality and
  complexity for codeword quantization.
  Therefore, it is desirable to develop high efficient descriptors, especially 
  for real-time applications, such as intelligent surveillance system with multiple 
  cameras, human-machine interaction and video games. 
  
  Gradient-based descriptors, on the other hand, are fast to compute. However, they 
  often show suboptimal performance and high feature dimensionality. 
Built on oriented gradient, the  HOG descriptor was originally introduced by Dalal and Triggs in \cite{Dalal2005} for human detection. It only contains local structure information and 
shows low recognition accuracy  on action recognition due to
  lack of motion information. For better performance, it is often combined
   with HOF  descriptor \cite{Laptev2008},  which requires the computation of dense optical flow.
 
 Kl{\"a}ser \cite{Klaeser} extended HOG descriptor from 2D image to spatio-temporal
 HOG3D descriptor with 3D oriented gradients. The spatio-temporal gradients 
 $(\frac{\partial I}{\partial x}, \frac{\partial I}{\partial y}, \frac{\partial I}{\partial t})$
      are computed for each pixel over the video, and saved in three integral videos. 
      Three gradient components of a ST patch can be computed efficiently from
      the integral videos. The mean 3D gradient vector from a local cell is quantized 
      using a regular polyhedron.  The 3D histograms of oriented gradients for the 3D
      patch are formed by concatenating gradient histograms of
      all cells.  Although it is very efficient to compute  gradients, 
       the  quantization with polyhedron for each sub-blocks is  expensive 
       considering the large number of patches sampled. The 3D quantization also results in high feature
       dimension. In addition, using  regular polyhedron with congruent faces to quantize
       the ST gradients may not be an optimal option because the units of spatial gradients and temporal 
       gradients in a video are different and should not be treated interchangeably.

Scovanner  \etal   \cite{Scovanner2007} extended 2D SIFT descriptor \cite{Lowe2003}
to represent spatio-temporal patches, called 3D SIFT.
Once gradient magnitude and orientation  computed in 3D,
each pixel  has two values $( \theta,  \phi )$ which
represent the direction of the gradient in three dimensions.
For orientation quantization,
the gradients  in spherical coordinates $( \theta,  \phi )$ are divided into 
 equally sized bins, which are represented by an $8 \times 4$ histogram. 
Such representation leads to singularity problems   as
bins get progressively smaller at the poles. Similar to HOG3D, it also has increased dimensionality,
and treats spatial  and temporal gradients as similar quantities.
 
In this paper, we focus on developing an efficient method for action recognition on realistic videos.
Our method is based on pure spatio-temporal gradients. However, our objectives are both to improve
the performance and to avoid high dimensionality of common 3D gradient-based descriptors, such as
HOG3D and 3D SIFT. In addition, we also aim at reducing the memory usage. To this end,
we made following main contributions:
\begin{itemize}
\item We propose a new spatio-temporal descriptor, called GBH,  which significantly outperforms 
other gradient-based descriptors both in performance and speed.
\item In an in-depth experimental evaluation, we demonstrate that  the recognition accuracy is preserved
while the spatial resolution is greatly reduced, which yields to both high efficiency and low
memory usage.
\item We rigorously show that  the proposed descriptor and its Fisher vector representation can
achieve real-time activity analysis with potential application in mobile computation.
\item We experimentally show that the GBH descriptor can improve recognition performance significantly 
when combined with HOF or MBH descriptor.
\end{itemize}
 
 The paper is organized as follows: The next section introduces
 the proposed GBH descriptor. 
 Section 3 details the methods we use. 
  Section 4 summarizes the evaluation methodology as well as the  datasets.
In section 5, 
we present some experimental results and analysis.
 we also provide the comparison of our methods with the state-of-the-art.
 The paper is completed with a brief conclusion.

\section{Gradient boundary histograms (GBH) descriptor}
\label{sect:gbh}

\begin{figure*}
\begin{center}
\includegraphics[trim = 3mm 38mm 6mm 16mm, clip, width=0.99\linewidth]{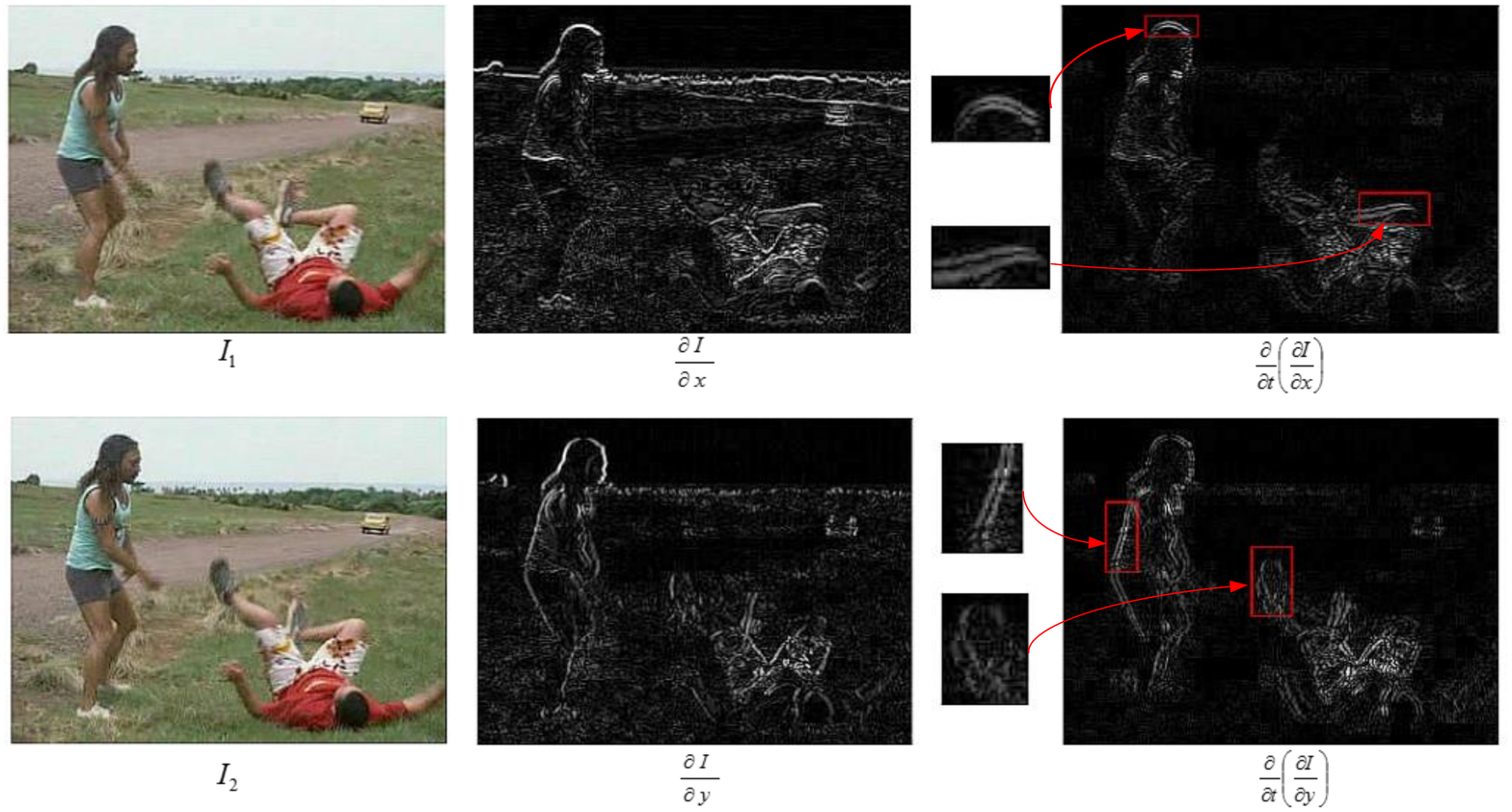}
\end{center}
   \caption{Illustration of gradients and gradient boundaries for a ``fall floor'' action.
      Compared to image gradients, gradient boundaries have less background noise. 
      More important, gradient boundaries encode motion information. The areas inside red bounding 
      boxes show the double edges at distances proportional to the speed of the moving
      body parts.}
\label{fig:model}
\end{figure*}

 In this paper, we  propose a new local spatio-temporal descriptor. 
 Our object is to avoid the expensive dense optical flow computation. We also intend to encode 
  both local static appearance  and motion information. However, we
want to avoid using three gradient components as 3D SIFT and HOG3D descriptor which 
lead to   high dimensionality and relatively expensive quantization cost. 
Instead, we adopt compact HOG-like descriptor with  two gradient components.
 
 For each frame in a video, we first compute image gradients using simple 1-D [-1,0,1] Sobel masks
 on both $x$ and $y$ directions. Then, we apply a [-1,  1] temporal filter over two consecutive 
 gradient images. Thus, for each pixel, we have:
 
 \begin{equation}
 I_{t,x} = \frac{\partial }{\partial t}(\frac{\partial I}{\partial x}), \ \ \ \ \ \   
  I_{t,y} =  \frac{\partial }{\partial t}(\frac{\partial I}{\partial y})
  \label{eq:gbhEq1}
 \end{equation}
  Now the gradient magnitude and orientation for each pixel are defined as follows:
 
  \begin{equation}
 r(x, y) = \sqrt{I_{t,x}^2 + I_{t,y}^2}, \ \ \ \ \ \ \ \  \   
   \theta(x,y) = \arctan(\frac{I_{t,y}}{I_{t,x}}) 
   \label{eq:gbhEq2}
  \end{equation}

Our new descriptor uses a histogram of orientation method, voting with $\theta$ and $r$
  as  in SIFT and HOG descriptors.
 However, instead of using image gradients, we use time-derivatives of image gradients, 
 which emphasize moving edge boundaries.
 We call this descriptor  gradient boundary histograms (GBH).
Figure~\ref{fig:model} illustrates the comparison of image gradients and gradient boundaries. 
We have two important observations here. First, 
 the subtraction of two consecutive image gradients results in the removal of 
 the backgrounds of the video sequences. The two gradient images in the centre show a lot of
 background noise, while the gradient boundary images on the right show clear human shapes
 with far less background noise. More important, gradient boundaries encode the moving human shapes. 
 As demonstrated by the red bounding  boxes in the figure, 
 the double edges at various distances are proportional to the moving speed of the human
    body parts.  For example, the distance between the leg double edges is larger than the 
  head double edges, because the leg moves faster than the head of the other person
   in the upper right image.

It seems  that the simple gradient subtraction works well only when the camera and background are largely static. 
However, our in-depth experimental evaluations show
that it also  achieves good performance on realistic HMDB51 dataset, which contains a lot of  camera motions and dynamic backgrounds. 
One possible explanation may be that 
the changes of the human gradient boundary (as shown in Figure~\ref{fig:model}) reflect the speed of the moving body parts.
Because the camera motion for two consecutive frames is often constant, the subtraction of two consecutive gradient images 
results in a constant  gradient offset. 
Therefore, the computed human gradient boundaries include the  absolute gradient displacements of human moving body
 plus a constant gradient displacement. Such constant gradient displacement from the camera motion has little performance impact.

\section{Video representation}
\label{sect:videoRep}
To demonstrate the performance of the proposed GBH descriptor, we apply an efficient 
random sampling scheme to extract the features, and use the state-of-the-art
local part model (LPM) \cite{Shi2013} to represent them. We finally use improved fisher vector \cite{Perronnin10} 
to encode features, followed by a linear SVM for classification.

\subsection{Review of LPM algorithm}

Local part model was introduced by Shi \etal in \cite{Shi2011}. Their original purpose was to address the
orderless issue of the bag-of-features representation by introducing overlapping local ``parts'' 
in the same spirit as bi-gram (n-Grams) \cite{Thurau2008}. In addition to having the overlapping
local part patches, the method also includes a coarse primitive level ``root'' patch which encodes
local global information.
To improve the efficiency of LPM computation, two integral videos are computed, one for the root
 at half resolution, and another one for the parts
at full resolution. The descriptor of a 3D patch can then be
computed very efficiently through 7 additions multiplied by
the total number of root and parts.

Later, Shi \etal \cite{Shi2013} improved the efficiency by combining random sampling
 method with local part model. Random sampling  does not require feature detection,
 which greatly improves processing speed.
In this work, the root and 8 parts are processed as
two separate channels. For each channel, a standard Fisher vector encoding 
is applied. The resulting Fisher vectors
from root and parts are concatenated
into one histogram for SVM classification.

The main challenge of applying FV encoding on the features of LPM is the high dimensionality.
The part channel in LPM contains a group of overlapping patches, and their histograms 
are concatenated into a high dimensional vector. 
Therefore, as discussed in next section, we 
reduce the descriptor dimensionality by  using Principal Component Analysis (PCA) 
on the purpose of better fitting FV encoding.
In our experiments, the dimensions of root
descriptor and part descriptor are reduced by 1/2 and 1/8, respectively.

\subsection{Fisher vector encoding} 
\label{sect:fv}

Recent studies \cite{wang2013b,Oneata2013,Sun2013} show that Fisher vector \cite{Jegou2012}
can improve performance over standard bag-of-features methods on action recognition.
  FV  extends the bag-of-features by
encoding high-order statistics  between the  descriptors and a Gaussian Mixture Model.
Since more information is encoded per visual word, 
 fewer visual words  are required than BoF, which makes FV more efficient to compute.

As indicated in \cite{Jegou2012},  it is  favourable to apply PCA dimensionality
reduction on feature vectors before FV encoding.
Moreover, for a $D$ dimension descriptor, the FV 
signature with $K$ words has an increased dimension of  $2DK$. 
In the case of local part model, the concatenated histograms of part patches result in
8 times the dimension of the used descriptor.
 Therefore we apply PCA on the computed LPM features.
 Our experimental evaluation shows that the feature dimensions can be reduced by 7/8 while preserving high
 accuracy. 

\section{Experimental setup}
In this section, we  introduce  implementation details of our evaluation
 methodology. We will also present the datasets and experimental parameters.

\subsection{Evaluation methodology} 
\label{sect:overview}

  As discussed in Section~\ref{sect:videoRep}, we use random sampling for feature extraction and
  local part model to represent the features. 
  We strictly follow the experimental settings as those in \cite{Shi2013}.
  However, we use Fisher vector encoding instead of bag-of-features. 
  
  We randomly choose 10000 features
  for each video with maximal video length of 160 frames.
  For those clips with more than 160 frames, we simply divide them into several segments,
  and sample features at same rate for each of them.
  
The sampled 3D patches are represented by GBH descriptor.
Under the LPM representation, each feature includes a root patch and a group of 
part patches (8 in our experiments).
Let $D_0$ be the feature dimension of the GBH descriptor. 
One LPM feature has two channels, with 1 root at dimension  $d_r=D_0$ and 8 parts 
of dimension  $d_p=8 \times D_0$.

For colour images, we simply choose the channel with largest
gradient values, which improves the accuracy by 0.2-0.5\%.

For Fisher vector encoding,
we use improved fisher vector \cite{Perronnin10} by applying the signed square-rooting 
followed by $L_2$ normalization, which
significantly improves the performance when combined with a linear classifier.
We set the number of visual words to $K = 128$
 and randomly sample  150,000 features from the training set 
to estimate the GMM and learn PCA projection matrix.
For Fisher vector encoding,  we first apply PCA to reduce  root vectors to 
 $d_r'=\frac{1}{2}d_r=\frac{1}{2}D_0$ and part
vectors  to $d_p'=\frac{1}{8}d_p=D_0$. 
Each video is, then, represented by a $2d_r'K+2d_p'K=3D_0K$ dimensional
Fisher vector.

The resulting Fisher vectors are fed into a linear
SVM implemented by LIBSVM \cite{Chang2011} with  $C=32.5$. 
For multi-class SVM,  we use one-against-rest approach.
To combine root and part channels of LPM representation, 
we simply concatenate the computed FVs from the respective channels.
The same strategy is used to combine multiple channels from different descriptors.

\subsection{Datasets} 
\label{sect:dataset}
 To demonstrate the performance and efficiency of the proposed descriptor, 
 we evaluate our method
 on two large-scale realistic action benchmarks, the UCF101 
  \cite{Soomro2012} and the HMDB51 \cite{Kuehne2011} datasets.
  
The \textbf{UCF101} dataset \cite{Soomro2012} is by far the largest human 
action dataset. It has 101 classes and 
13320 realistic video clips extracted from YouTube.
All clips are encoded at a resolution of $320\times240$ and at a
frame rate  of 25 FPS.
 The clips of one action class are divided
into 25 groups. The dataset is very large and relatively challenging.
We report average accuracy over three distinct training and testing splits 
as proposed in \cite{Soomro2012}.
For split 1, split 2 and split 3,  clips from
groups 1-7, groups 8-14 and groups 15-21 are selected respectively as testing samples,
 and the rest for training.

The \textbf{HMDB51} dataset \cite{Kuehne2011} contains 51 action categories, 
with a total of 6,766 video clips extracted from various sources, such as
  Movies, the Prelinger archive, Internet, Youtube and Google videos.
 It is perhaps  the most  challenging dataset with realistic settings. 
 The videos have different aspect ratio, but with a fixed height of 240 pixels.
The clips have various video quality, and the minimum quality standard
is at  60 pixels in height for the main actor. 
We use the original non-stabilized videos with the same three train-test splits 
 \cite{Kuehne2011}, and report the average accuracy over the three splits 
in all experiments.

\begin{center}
 \begin{table*}[!tpb]
\begin{center}
\begin{tabular}{|c|| c|c|c|| c|c|c|}
\hline			
 \multirow{2}{*}{Smoothing} &\multicolumn{3}{c||}{HMDB51 }&\multicolumn{3}{c|}{UCF101 }\\
  &HOG& HOG3D & GBH &HOG& HOG3D & GBH \\ 
  \hline
Yes &29.4\%$\pm0.6$ & 37.8\%$\pm0.6$ &  44.4\%$\pm0.3$ &60.6\%$\pm0.2$ & 64.5\%$\pm0.9$ &  74.6\%$\pm0.3$ \\
  No & 30.0\%$\pm0.3$ &38.2\%$\pm0.3$ &  40.2\%$\pm0.7$ &61.2\%$\pm0.5$ & 64.7\%$\pm0.4$ &  73.0\%$\pm0.6$  \\ 
   
\hline 
\end{tabular}
\end{center}
\caption{The performance impact of Gaussian smoothing on different descriptors. The 
experiments are performed on the video at original resolution. }
\label{tab:gaussSmooth}
 \end{table*}
\end{center}

\begin{center}
 \begin{table*}[!tpb]
\begin{center}
\begin{tabular}{|r|| c|c|c|| c|c|c|}
\hline			
 \multirow{2}{*}{Resolution} &\multicolumn{3}{c||}{HMDB51 }&\multicolumn{3}{c|}{UCF101 }\\
  &HOG& HOG3D & GBH &HOG& HOG3D & GBH \\ 
  \hline
(avg.)364 x 240 &30.0\%$\pm0.3$ &38.2\%$\pm0.3$ &  44.4\%$\pm0.3$ &61.2\%$\pm0.5$ & 64.7\%$\pm0.4$ &  74.6\%$\pm0.3$ \\
  182 x 120  & 27.5\%$\pm0.5$ &36.5\%$\pm0.2$ &  44.7\%$\pm0.3$ &55.4\%$\pm0.4$ & 63.6\%$\pm0.2$ &  74.2\%$\pm0.4$  \\ 
    91 x 60 &23.7\%$\pm0.4$ & 33.0\%$\pm0.7$ & 45.3\%$\pm2.4$ &50.5\%$\pm0.4$ & 56.6\%$\pm0.7$ &  73.6\%$\pm0.9$ \\
\hline 
\end{tabular}
\end{center}
\caption{The performance comparison of three gradient-based descriptors at
  different spatial resolutions.
 }
\label{tab:numFeature}
 \end{table*}
\end{center} 

\subsection{Parameters} 
\label{sect:para}
In this section, we present our  parameter settings, which determine the features' dimension.
We test and compare our GBH descriptor with other descriptors.
However,  we use the simplified HOG3D, HOG and MBH
descriptors with reduced  dimensionality, mainly by controlling the  number of
cells per ST patch.

\textit{Notation:} we define the sampling grid at
half  the spatial resolution of the processed video. The root patches are randomly
chosen from this half size video, and we will refer to it as ``root video''.
The part patches are sampled from the processed video at full spatial resolution, 
which is referred to  as ``part video'' or ``processed video'', interchangeably.
We also use ``original video'' to represent the original  spatial resolution of the 
clips from the datasets. 

\textbf{Random sampling.}
In order to provide comparable results, 
we strictly follow the sampling parameter settings as those in \cite{Shi2013}.
We first define a very dense sampling grid over the root video. 
 A 3D video patch centred at $\mathit{(x, y, t)}$  is sampled 
 with a patch size determined by the multi-scale  factor $\mathit{(\sigma,\tau)}$.
  The consecutive scales are computed by multiplying $\mathit{\sigma}$
 and $\mathit{\tau}$ by a factor of $\sqrt{2}$. In our experiments,
we  set  minimal temporal size to 14 frames, and choose the optimal minimal spatial size 
based on the descriptors and the size of the processed video.
The sampling step size is determined by  multiplying patch size by a factor of 0.2.
  With a total of 8 spatial scales and 2 temporal scales, 
 we sample a video 16 times. 

We randomly sample 10000 root patches from the root video.
For each root patch, we sample 8 ($2 \times 2 \times 2$) overlapping part patches
from the part video.The histograms of 1 root patch and 8 part patches
are treated as two separate channels.

\textbf{ GBH}, \textbf{ HOF}, \textbf{ MBHx} and \textbf{ MBHy}.
 Each patch is subdivided into a grid of $2\times2\times2$ cells, 
 with no sub-block division. 8 bins are used for quantization, which leads to
 a feature dimension of 64. Thus,  a LPM feature
 has a root channel of dimension 64 and a part channel of  dimension 512.
 
 We evaluate GBH descriptor on the processed video with different spatial resolutions.
When the processed video has same size as original video, the initial patch size is  $28\times28\times14$.
 We use minimal patch size of $20\times20\times14$ and $10\times10\times14$, respectively,
  for the part video size  at half and one quarter.

  \textbf{HOG}. 
 For HOG, we use same parameters as GBH in most cases.
 However, for original video size, 
  the optimal minimal patch size is $24\times24\times14$, 
 and each patch is subdivided into a grid of $2\times2\times2$ cells, 
  with $2\times2\times2$ sub-block divisions. With 8 bins quantization,
It has same feature dimension as GBH.

\textbf{HOG3D}. 
The HOG3D parameters are: number of histogram cells $M = 2$, 
$N = 2$; number of sub-blocks $1\times1\times3$;  and polyhedron type dodecahedron(12)  with full orientation.  
The optimal minimal patch size is $24\times24\times14$ for original video size.
 With one HOG3D descriptor at dimension
 of 96 ($2\times2\times2\times12$), our local part model feature 
has a dimension of  96  for the root channel and 768 for the part channel.

\section{Experimental results}

In this section, we evaluate performance of the proposed GBH descriptor
 for action classification on two realistic  datasets.
 Due to the random sampling, we repeat the experiments 3 times, and report 
mean accuracy and standard deviation over 3 runs.

\subsection{Influence of Gaussian smoothing}
\label{sect:smooth}
We first  evaluate the impact of the Gaussian smoothing. 
The results are show in Table~\ref{tab:gaussSmooth}.
All the experiments are performed on original video at full resolution. 
If smoothing is ``Yes'', a Gaussian filter is applied on all frames before computing 
the gradients. 

For HOG and HOG3D descriptors, we observe  slight performance drops on all cases
when applying Gaussian filter before computing gradients. Similar performance drop is
reported on HOG  on human detection \cite{Dalal2005} with smoothing. 
The performance of GBH, on the other hand, increases significantly by pre-smoothing, 
with 4.2\% on HMDB51 and 1.6\% on UCF101.
Such a performance increase may be explained by the fact that the second order 
derivatives are more sensitive to noise. In addition,
applying Gaussian smoothing before gradient subtraction results in the suppression of certain
background gradients, such as  tree leaves and grass textures (as shown in Figure~\ref{fig:model}).
Such background  textures are often a huge challenge  to optical flow estimation. 

 \begin{center}
 \begin{table*}[tpb]
 {\small
 \hfill{}
 \begin{center}

\begin{tabular}{|l||c|c|c|c|c|c|c|c|}
\hline			
  Dataset&GBH& HOG &HOG3D & HOF& MBH &HOG+HOF& HOF+GBH&  MBH+GBH\\ 
  \hline
 HMDB51 &   44.7\% & 30.0\%  &  38.2\% &   39.9\%  & 54.7\% &  45.6\% &   51.3\%  &  58.8\% \\ 
UCF101 &  74.2\%  &   61.2\% &   64.7\%  &    65.9\% &81.0\%   &   75.4\% &  78.0\% & 84.0\%  \\
\hline 
\end{tabular}
\end{center} }
\hfill{}
 \caption{Performance comparison of the proposed GBH descriptor and other local descriptors. }
 \label{tab:GBHcompar}
 \end{table*}
\end{center} 
 
\subsection{Evaluation of GBH descriptor}

Table~\ref{tab:numFeature} shows the performance comparison of three gradient-based 
descriptors in  different spatial resolutions.
For HOG and HOG3D descriptors, the performance  is
consistently and significantly decreased for both HMDB51 and UCF101 when
the spatial resolution is reduced. Such results are consistent with observations 
in \cite{Wang2009} on Hollywood2 dataset. As resolution is reduced, the
background gradients interfere with the gradients associated with human subjects.

For GBH descriptor, one very important observation  is that the accuracy is preserved 
on HMDB51 and with little (1\%) loss on UCF101
when the spatial resolution is reduced by a factor of $\lambda=4$.
This leads to huge benefits in efficiency considering that the sub-sampling in resolution 
by  $\lambda$ results in a reduction by a factor of $\lambda^2$ on both number of processed pixels
and memory usage.
Kl{\"a}ser \etal \cite{Klaeser} observed, in a HOG3D approach, that using integral video can result in 
a reduction factor of  $21$  in memory usage when compare to spatio-temporal
``pyramids''. Our method uses even less memory  when processing video at low
resolution.

When processing video at a very low resolution, we observe a relatively high standard deviation 
on performance for both HMDB51 (2.4) and UCF101 (0.9). This is probably due to the fact that
the sampling is performed on the very low resolution video. At such low resolution, a 
sampled ST patch could have large differences even for a one pixel displacement.
At this point, it is unclear why the GBH descriptor performs better at very low 
spatial resolution (91 x 60) on HMDB51 than on UCF101.
 Our hypothesis is that  high sampling density on clips with fewer frames may include
more information, which could provide bias benefits for the short clips of HMDB51.
 Moreover,  the  HMDB51 has a quality standard 
 of a minimum of 60 pixels in height for the main actor, which may improve the robustness 
 when the spatial resolution is greatly reduced.

\subsection{Comparison of GBH  and other descriptors} 
We perform a number of experiments to evaluate our proposed GBH descriptor.
Table~\ref{tab:GBHcompar} shows the performance comparison of the  GBH descriptor and other local descriptors. 
The evaluation is performed under a common experimental setup. 
The HOG and HOG3D descriptors  are computed  at full resolution, and 
other descriptors  are computed at half the resolution.
We use the default parameters as in Section~\ref{sect:para}, and randomly choose
10K features (that is 10K root patches + 80K part patches) from each clip with up to 160 frames.
The dense optical flow for HOF and MBH descriptors is computed using efficient  Farneb\"{a}ck's approach \cite{Farnebaeck03}.

The GBH descriptor gives surprisingly good results by itself, with $44.7\%$ on HMDB51 and $74.2\%$
on UCF101. It outperforms HOG, HOG3D and HOF descriptors on both datasets.
 However, the MBH descriptor outperforms all other descriptors by a large margin. 
 When combined with flow-based HOF or MBH descriptor, we observe significant performance improvements.

\NewDocumentCommand{\rot}{O{90} O{1em} m}{\makebox[#2][l]{\rotatebox{#1}{#3}}}%
\begin{table}
 \begin{center}
\begin{tabular}{|l|l||c|c|}
\hline			
\multicolumn{2}{|l||}{Method } & HMDB51  & UCF101 \\ 
  \hline
  \multicolumn{2}{|l||}{HMDB51 \cite{Kuehne2011} } & 23.2\% &  --   \\  
 \multicolumn{2}{|l||}{UCF101 \cite{Soomro2012} } & -- &  43.9\% \\ 
  \multicolumn{2}{|l||}{Efficient OF \cite{Kantorov14}  } & 46.7\%$^{*}$ & --   \\ 
  \multicolumn{2}{|l||}{DCS \cite{Jain13} } &  52.1\%$^{*}$  &  --  \\ 
   \multicolumn{2}{|l||}{FV coding\cite{Oneata2013} } & 54.8\%$^{*}$  & --   \\ 
 \multicolumn{2}{|l||}{Trajectories \cite{wang2013b} } & 57.2\%$^{*}$  &   85.9\%$^{*}$   \\ 
 \hline 
 \multirow{5}{*}{\rot{Ours} }
 &$\mathrm{GBH}$ & 44.7\%$\pm0.3$& 74.2\%$\pm0.4$ \\  
  &$\mathrm{MBH^1}$ & 54.7\%$\pm0.2$ & 81.0\%$\pm0.2$ \\ 
   &$\mathrm{MBH^2}$ & 58.9\%$\pm0.3$ & 84.7\%$\pm0.1$ \\ 
\hhline{~---} &$\mathrm{GBH+MBH^1}$ & 58.8\%$\pm0.2^{*}$ &84.0\%$\pm0.2^{*}$  \\ 
&$\mathrm{GBH+MBH^2}$ & \textbf{62.0}\%$\pm0.2^{*}$ &\textbf{86.6}\%$\pm0.2^{*}$  \\ 
\hline 
\end{tabular}
 \end{center}
\caption{Comparison of average accuracy on HMDB51 and UCF101  with state-of-the-art methods. Those  marked with $^{*}$ are results with multiple descriptors.
The $\mathrm{MBH^1}$ is based on the Farneb\"{a}ck's optical flow method \cite{Farnebaeck03},
and $\mathrm{MBH^2}$ uses duality-based TV\_L1 approach \cite{Zach2007}. }
\label{tab:compare1}
\end{table}
\subsection{Comparison to state-of-the-art} 

Table~\ref{tab:compare1} shows the comparison of our method with the state-of-the-art.
We set the part video at the half the spatial resolution of original video, 
and use the parameters listed in Section~\ref{sect:para}.
Most state-of-the-art methods use multiple descriptors and apply some feature encoding
 algorithms to improve the performance.  For example, Jain \etal \cite{Jain13} 
 combine five compensated descriptors and apply VLAD representation. 
Wang and Schmid \cite{wang2013b} use four descriptors and  Fisher Vector encoding.
They also improve the performance with human detection and extensive camera motion compensation.

 On \textbf{HMDB51}, our method achieves   $62.0\%$ when combining GBH with MBH descriptors, 
 which outperforms the state-of-the-art result ($57.2\%$ \cite{wang2013b}) by $5\%$.
With MBH computed from efficient  Farneb\"{a}ck optical flow method, we obtain $58.8\%$ on 
the two descriptors (MBH+GBH), which also  exceeds state-of-the-art results. 
Note that our results are obtained from two descriptors combined with Fisher vector.
Since we have both root and part channels, we set  $K = 128$ for FV encoding.
Thus, our resulting FV features have same dimensionality as those in  \cite{wang2013b},
which uses $K = 256$. In the case of $K = 256$,  we achieve $\textbf{63.2\%}$.

 On \textbf{UCF101},  we report  $86.6\%$ with two descriptors.
  It slightly outperform the state-of-the-art result \cite{wang2013b} ($85.9\%$), which is obtained
 with four descriptors and  Fisher Vector encoding as well as extensive camera motion estimation.

 \begin{center}
 \begin{table*}[tpb]
 {\small
 \hfill{}
 \begin{center}
\begin{tabular}{|c||c| c|c|c|c|c|c|c|c|}
\hline	
 \multirow{2}{*}{Dataset}   & \multirow{2}{*}{Resolution}   &\multicolumn{4}{c|}{Speed   (frames per second) }&\multirow{2}{*}{Mean accuracy}\\
  	\hhline{~~~~~~~~}	 & &Integral video&  Sampling& FV encoding & Total fps& \\  						
   	\hline 			
   \multirow{3}{*}{HMDB51}   & \multirow{1}{*}{364 x 240} &120.5&249.5&151.7&52.8&$44.4\%$\\ 	  									
     	\hhline{~~~~~~~~}	  & \multirow{1}{*}{182 x 120} &302.9&261.4&150.5&72.5&$44.7\%$\\ 
    	\hhline{~~~~~~~~}	  & \multirow{1}{*}{91 x 60} &437.5&294.4&156.8&82.7&$45.3\%$\\ 			
\hline  
  \multirow{3}{*}{UCF101}   & \multirow{1}{*}{320 x 240}&132.8&356.7&220.2&67.2&$74.6\%$\\ 	  									
   	\hhline{~~~~~~~~}	  & \multirow{1}{*}{160 x 120}&362.9&384.7&223.7&101.8&$74.2\%$\\ 
  	\hhline{~~~~~~~~}	  & \multirow{1}{*}{80 x 60}&596.2&422.2&228.5&118.5&$73.6\%$\\ 			
\hline 
\end{tabular}
\end{center} }
\hfill{}
 \caption{Average computation speed with single core at different stages in frames per second. $K = 128$ codewords per channel is used for FV encoding, and 10K features are sampled in the experiment. 
 The dimensionality reduction process is included in FV encoding. Note that
 the classification stage is not included. }
 \label{tab:compute}
 \end{table*}
\end{center} 

\begin{center}
 \begin{table*}[tpb]
 {\small
 \hfill{}
 \begin{center}
\begin{tabular}{|c||c| c|c|c|c|c|c|c|c|}
\hline	
 \multirow{2}{*}{Sampling \#}   & \multirow{2}{*}{Resolution}   &\multicolumn{4}{c|}{Speed   (frames per second) }&\multirow{2}{*}{Accuracy}\\
  	\hhline{~~~~~~~~}	 & &Integral video&  Sampling& FV encoding & Total fps& \\

   	\hline 												
     	\hhline{~~~~~~~~}4K  & \multirow{1}{*}{182 x 120} &52.7&267.6&89.3&29.0&$43.3\%$\\ 			
\hline    									
   	\hhline{~~~~~~~~}	10K  & \multirow{1}{*}{182 x 120}&52.9&108.4&37.5&18.3&$44.7\%$\\ 		
\hline 
\end{tabular}
\end{center} }
\hfill{}
 \caption{Average computation speed on a Toshiba Netbook with an AMD-E350 cpu and 2GB memory. 
 The experiments are performed on HMDB51 dataset.
 $K = 128$ codewords per channel is used for FV encoding. 4K and 10K features are 
 sampled  in two different experiments. The dimensionality reduction process 
 is included in the FV encoding.  }
 \label{tab:compute2}
 \end{table*}
\end{center}
\subsection{Computational efficiency}
We perform a number of experiments to evaluate the efficiency of GBH descriptor.
We use VLFeat library \cite{vedaldi08vlfeat} for Fisher vector encoding.
The runtime is estimated on an Intel i7-3770K PC with 8GB memory. 
The prototype is implemented in C++.
In order to avoid built-in multi-core processing of  VLFeat library and OpenCV library, 
we set only one core active @ 3.5Ghz in Bios, and disable both Hyper-threading and Turbo-boost. 
For all experiments,  $K = 128$ visual words per channel are used for Fisher vector encoding.

Table~\ref{tab:compute} compares the computational speed under different spatial resolutions.
For all cases, GBH achieves high processing speed, reported in 
frames per second. Using low resolution results in
an impressive total fps  at only little performance cost.
There is little speed difference
between HMDB51 and UCF101, mainly because on average we sample 10K features per 95 frames on
HMDB51, but per 160 frames on UCF101.

Table~\ref{tab:compute2} lists average computation speed on  an AMD-E350 CPU, which
also shows a high processing frame rate. This proves the high efficiency of GBH descriptor,
 and demonstrates its potential for real-time applications as well as mobile recognition.

\section{Discussion}
The GBH descriptor shows good performance with high efficiency.
One possible application is action recognition for surveillance system with multiple 
static cameras.
The efficient action detection, which involves the 
evaluation of multiple detection windows, can also benefit from the approach.
GBH could also be used to improve the HOG performance on human detection from the video.
To further improve performance, we can also explore deeper part hierarchies (\ie parts with parts).
Since the GBH performance is preserved when using low resolution,
we could add another layer of parts on full resolution videos.
Our preliminary experiments on GBH descriptor with FV show that
adding another level of parts  improves the performance from 44.7\% to 46.0\% on HMDB51,
and  from 74.2\% to 76.1\% on UCF101. \par
We have shown that GBH can be combined with  flow-based HOF or MBH descriptors to improve its performance.
The dense optical flow on which HOF and MBH rely  is relatively expensive to compute. However,
this optical flow can be estimated very efficiently in compressed domain \cite{yeo2006}.
Recent progress in hardware and network technologies leads to motion-encoded videos, 
such as those obtained from internet protocol (IP) cameras in security and surveillance video systems.
A recent study \cite{Kantorov14} also demonstrates the  efficiency of extracting optical flow in
 compressed video, and achieves  results similar to our GBH descriptor (45.4\% on HMDB51 with MBH).
  Nevertheless, such  approaches could be combined with our 
GBH descriptor to achieve state-of-the-art performance with high efficiency.\par
We have demonstrated that GBH outperforms other gradient-based HOG, HOG3D descriptors and flow-based
HOF descriptors. Also, the GBH descriptor works well not only on
static videos but also on non-static-camera videos, such as HMDB51. 
The performance may be further improved by explicitly estimating camera motion \cite{wang2013b}
or combining it with efficient MBH descriptor \cite{Kantorov14}. The main purpose of the GBH descriptor 
is to improve efficiency as well as to provide rich motion information.
An efficient descriptor with high performance 
can find many different applications in video analysis. \par
The main challenge in using FV is the resulting high dimensionality,
which is expensive  in classification stage even by using linear SVM.
In comparison with BoF (often with 4K words), 
the  FV representation is not much as a ``Compact Feature Set'' as the claim  made in \cite{Oneata2013}.
The original FV approach \cite{Jegou2012}, however, reduces the high dimensional FV into a compact low 
dimensional vector and observes improved performance on image search.
Considering its high dimensionality, it is desirable to improve the performance of individual 
descriptor, and combine fewer descriptors for high accuracy.

\section{Conclusions}

This paper introduces a spatio-temporal descriptor for action recognition. 
The proposed descriptor is based on ST gradients, and outperforms other 
gradient-based descriptors. We demonstrate its benefits in combination with 
Fisher vector representation.
We experimentally show that the performance is preserved
even when the spatial resolution is greatly reduced.
Compared with existing methods, a major strength of our method
is its very high computational efficiency, with potential for mobile applications.

{\small
\bibliographystyle{ieee}
\bibliography{egbib}
}

\end{document}